%% file: icme2023template.tex
\documentclass{article}
\usepackage{spconf,amsmath,epsfig, graphicx,enumitem,amsfonts, makecell}

\let\OLDthebibliography\thebibliography
\renewcommand\thebibliography[1]{
  \OLDthebibliography{#1}
  \setlength{\parskip}{0pt}
  \setlength{\itemsep}{0pt plus 0.3ex}
}

\begin{document}\sloppy

\def\x{{\mathbf x}}
\def\L{{\cal L}}

\title{AFN:~Adaptive Fusion Normalisation via an Encoder-Decoder Framework}
%
\name{Anonymous ICME submission}
\name{Zikai Zhou, Shuo Zhang, Ziruo Wang, Huanran Chen}
\address{}

\maketitle

\begin{abstract}
Normalisation is crucial for high-performing machine learning models, especially  deep neural networks. A plethora of normalisation functions has been proposed but they  were designed for specific purposes and thus are not general to various application scenarios. In response, efforts have been made to design a unified normalisation function that combines normalisation procedures and mitigates their weaknesses. In this paper, we propose a novel normalisation function called Adaptive Fusion Normalisation (AFN). Through experiments, we demonstrate that AFN outperforms previous normalisation techniques in domain generalisation and image classification tasks.
\end{abstract}
\begin{keywords}
Adaptive Fusion Normalisation, Domain Generalisation, Image Classification.
\end{keywords}

\input{sections/intro}

\input{sections/related}

\input{sections/methodology}

\input{sections/exp}

\input{sections/conclusion}

\bibliographystyle{IEEEbib}
\bibliography{icme2023template}

\end{document}

%% file: sections/intro.tex
\section{Introduction}
\label{sec:intro}

Normalisation layers have played a crucial role in the remarkable success of deep learning. Most of the state-of-the-art models contain one or more normalisation layers, which normalise mainly by mean and variance in each sequence or batch to make the distribution of the input of each layer approach Gaussian distribution, streamline the training of neural networks to work, and prevent gradient vanishing or gradient explosion at the same time.

\begin{figure}[t]
    \centering
    \includegraphics[width=1\linewidth]{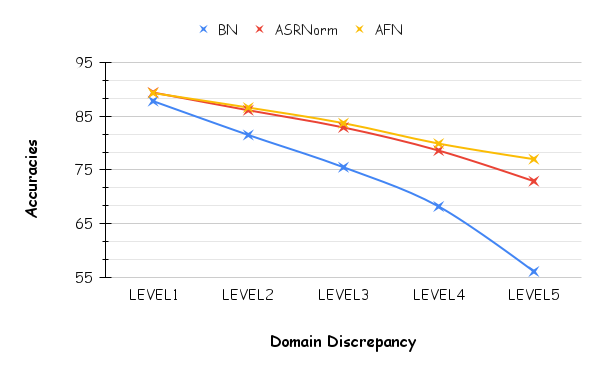}
    \caption{Accuracy of three different normalisation methods for single domain generalisation on \textbf{CIFAR-10-C}, compared at five different levels of domain discrepancy brought by corruptions. }
    \label{cifar10c}
\end{figure}

Various normalisation techniques have been proposed, each with its own set of advantages and disadvantages. Batch Normalisation~\cite{BatchNorm} (BN) normalises the input information by calculating the mean and variance value between mini-batch, effectively combating overfitting, but leads to training-testing in-consistency. Layer Normalisation~\cite{LayerNorm} (LN) addresses this issue by utilising mean and variance across different dimensions of the input, aligning training and testing. 
Instance Normalisation~\cite{Ulyanov2016InstanceNT}(IN) normalises the activation of individual instances within a batch, accelerating training convergence, but leads to instability in training, particularly with small batches. Nonetheless, it typically lags behind BN in computer vision tasks. Group Normalisation~\cite{GroupNorm} (GN) divides features into several groups and normalises them by statistics in each group, whose performance is less dependent on the batch size compared with BN, but has some disadvantages such as being sensitive to distortion or noise introduced by regularisation~\cite{GroupNorm} such as Dropout~\cite{srivastava2014dropout}. The strengths and weaknesses of these techniques have motivated researchers to explore mixed approaches that leverage their benefits.

By adding a few parameters, Switchable Normalisation~\cite{SwitchNorm} (SN) unifies BN, IN and LN. By using the gate parameters, Batch-Instance Normalisation~\cite{Nam2018BatchInstanceNF} (BIN) combines the advantages of BN and IN, which can both improve the performance on BN-based image classification tasks, and IN-based image style transfer tasks. Adaptive Scale and Rescale Normalisation~\cite{ASRNorm} (ASRNorm) adds more parameters to normalisation layers and unifies BN, LN, GN, and SN, and combines them with Adversarial Domain Augmentation~\cite{ADA,MADA,MEADA}, achieving the state-of-the-art result in many applications of domain generalisation. However, we observe that transitioning from other normalisation layers to ASRNorm can occasionally lead to gradient instability, resulting in poor performance.

In order to make those kinds of normalisation layers more suitable for different tasks, we design a new normalisation function denoted as Adaptive Fusion Normalisation(AFN), which combines the structure of ASRNorm and BN. Also, we design hyper-parameters to make our normalisation more similar to BN at the earlier parts of training while having a better performance in the later parts of training, which can easily adapt to the data.

Our contribution can be summarised as follows: 
\begin{itemize}
  \item We design a new normalisation layer called AFN by combining the structure of BN and ASRNorm to make it more suitable for image classification task.
  \item We carry out extensive experiments to show that our normalisation layer has taken advantage of BN and ASRNorm and outperforms them in domain generalisation and image classification.
\end{itemize}

%% file: sections/related.tex
\section{Related Work}
\textbf{Adaptive Scale and Rescale Normalisation.} By adding additional parameters to IN, ASRNorm~\cite{ASRNorm} shows a great improvement on Adversarial Data Augmentation~\cite{ADA, MADA, MEADA, chen2023rethinking, huang2023t}. However, we observed that pretraining a model with other normalisation layers and switching it by ASRNorm sometimes leads to gradient explosions and bad outcomes. Moreover, training a model from scratch solely with ASRNorm, without the assistance of Adversarial Domain Augmentation, often leads to unsatisfactory performance.

%% file: sections/methodology.tex
\section{Methodology}
\begin{figure*}[ht]
\centering
\includegraphics[width=1\linewidth]{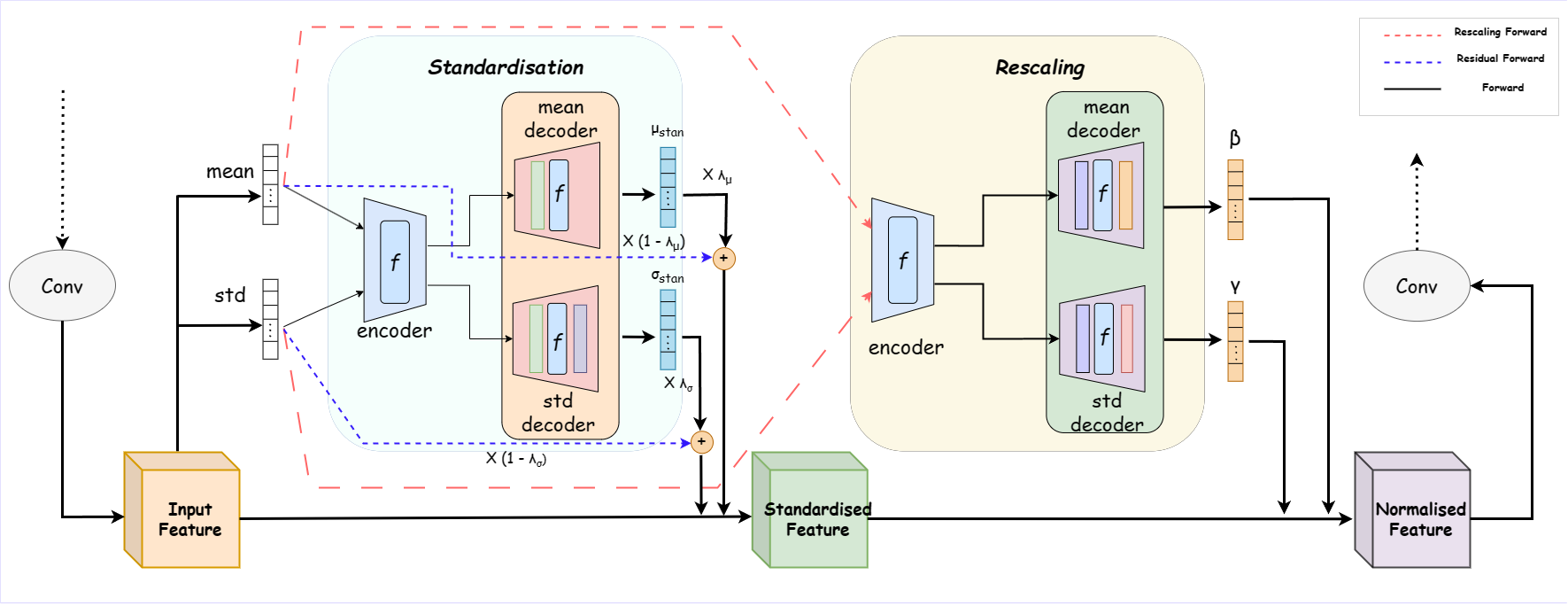}
\caption{Overview of our method, which is divided into two steps: standardisation and rescaling. We use an encoder-decoder framework to learn both the standardisation and rescaling statistics from the mini-batch statistics of the input. For the standardisation stage, we use a residual learning framework to render our training process stable.}
\label{method}
\end{figure*}

The main distinction between our method and ASRNorm lies in the normalisation approach. Our method employs statistics computed between each batch, whereas ASRNorm utilises statistics computed between each instance. Consequently, our approach can be viewed as an extension of BN with added parameters, while ASRNorm can be seen as augmenting IN with additional parameters. Moreover, our method outperforms ASRNorm in domain generalisation tasks, and can also be applied to image classification tasks, whereas ASRNorm cannot. 
In the image classification task, our method outperforms previous normalisation methods. Figure \ref{method} is the overview of our method, dividing into standardisation and rescaling parts.

\subsection{Standardisation Process}

Consider an input feature map with shape $(N, C, H, D)$, where $N$ is the batch size, $C$ is the number of channels and $H, D$ is the width and height of the feature map, respectively. First, we reshape the feature map to $(N \cdot H \cdot D, C)$, just as in the BN using statistics from the dimension $(N, H, D)$ instead of merely from $(N)$ to be more efficient without decreasing the performance. Then, we compute the mean and variance of this input batch by using each sample $x_i$ from $(N \cdot H \cdot D, C)$ dimension as:

\begin{equation}
    \mu = \frac{\sum_{i=1}^{N\cdot H\cdot D} x_i}{N\cdot H\cdot D}, \; \sigma =\sqrt{ \frac{\sum_{i=1}^{N\cdot H\cdot D} (x_i-\mu)^2}{N\cdot H\cdot D}}.
\end{equation}

Subsequently, we follow the setting of ASRNorm, which uses an encoder-decoder structure, whose bottleneck ratio is set to 16, to force our neural network to provide more suitable statistics for this mini-batch. The encoder extracts global information by interacting with the information and the decoder learns to decompose the information. For efficiency, both the encoder and decoder consist of one fully connected layer. ReLU~\cite{Nair2010RectifiedLU} is used to render the feature non-linear and ensure that $\sigma_{stan}$ is non-negative:

\begin{equation}
\centering
    \begin{aligned}
        \mu_{stan} = f_{dec}(ReLU(f_{enc}(\mu))) \\
    \sigma_{stan} = ReLU(g_{dec}(ReLU(g_{enc}(\sigma)))).
    \end{aligned}
\end{equation}
The encoders project the input onto the hidden space $\mathbb{R}^{C_{stan}}$, while the decoders project it back onto the space $\mathbb{R}^C$, where $C_{stan} < C$, and  $f_{enc} = g_{enc}$.

Then, we linearly combine the original statistics and statistics from our neural network, using the residual learning framework to make our training process stable.
 In the early training stages, the scale of $\hat{x}$ can be large due to the scale of input $x$, potentially leading to gradient explosion. To balance the scale between $\mu_{stan}$ and $\sigma_{stan}$, we introduce a residual term for regularisation.

\begin{equation}
\centering
    \begin{aligned}
    \label{formular:residual}
    \hat{\mu}_{stan} = \lambda_\mu * \mu_{stan} + (1-\lambda_\mu)*\mu \\
        \hat{\sigma}_{stan} = \lambda_\sigma * \sigma_{stan} + (1-\lambda_\sigma)*\sigma,
    \end{aligned}
\end{equation}
 where $\lambda_\mu$ and $\lambda_\sigma$ are learnable parameters ranging from 0 to 1 (bounded by sigmoid). Because the neural network cannot give a good statistic at the beginning of training, we initialise $\lambda_\mu$ and $\lambda_\sigma$ into $sigmoid(-3)$ to make $\lambda_\mu$ and $\lambda_\sigma$ approximate 0. Smaller than $sigmoid(-3)$ values will lead to a gradient vanishing problem, which is not desirable at training time.

Finally, we will get our normalised features by:
\begin{align}
\label{bar}
    \bar{x} = \frac{x-\hat{\mu}_{stan}}{\hat{\sigma}_{stan}}.
\end{align}

\subsection{Rescaling Process}
BN utilises two additional parameters to rescale the normalised features. Just like ASRNorm, we also use two additional neural networks to compute the weights and bias for the rescaling process.

\begin{equation}
\centering
\begin{aligned}
    \hat{\gamma} = \lambda_\gamma sigmoid(\phi_{dec}(ReLU(\phi_{enc}(\sigma)))) + \gamma_{bias} \\
    \hat{\beta} = \lambda_\beta tanh(\psi_{dec}(ReLU(\psi_{enc}(\mu)))) + \beta_{bias},
\end{aligned}
\end{equation}
where $\beta_{bias}$ and $\gamma_{bias}$ are two learnable parameters identical to those of BN, and $\lambda_\beta, \lambda_\gamma$ are learnable parameters, which are initialised with a small value $sigmoid(-5)$, to smoothen the learning process. We resume the $\beta_{bias}$ and $\gamma_{bias}$ from the BN layers in the pre-trained model, so each stage of the neural network acts like an identity function at the beginning of the training. 
Also, encoders and decoders are fully connected layers, and sigmoid and tanh functions are used to ensure the rescaling statistics are bound. The encoders project the inputs onto the hidden space $\mathbb{R}^{C_{rescale}}$ with $C_{rescale} < C$. The decoders project the encoded feature back onto the space $\mathbb{R}^C$.

Next, we rescale the normalised feature just like in any other normalisation, and send it to the next module of the neural network:
\begin{align}
    \hat{x} = \bar{x} * \hat{\gamma} + \hat{\beta}.
\end{align}

\subsection{Module Inability}

\begin{figure}[t]
    \centering
    \includegraphics[width=1\linewidth]{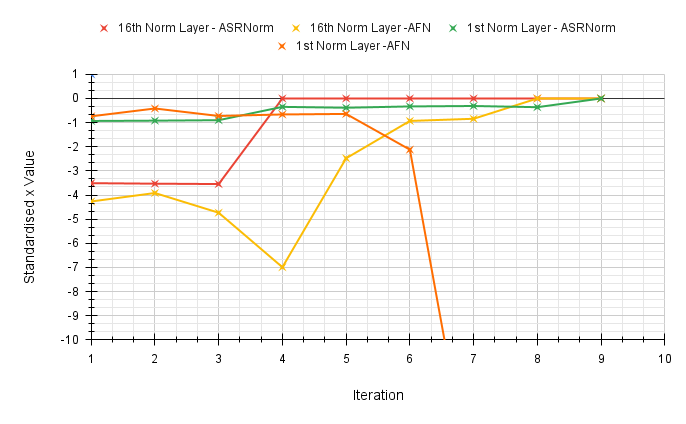}
    \caption{Input flow in VGG19\_BN on SVHN dataset. VGG19\_BN has 5 blocks, totally 16 normalisation layer. We choose 1st and 16th normlisation layers to show the gradient explosion/vanishing in AFN/ASRNorm.}
    \label{inability}
\end{figure}

As shown in Fig \ref{inability}, we take VGG19\_BN architecture as an example to illustrate ASRNorm has gradient explosion/vanishing problems. With the increase of iteration, $\bar{x}$ from Eqn \ref{bar} is gradually approaching 0, leading to gradient vanishing. On the contrary, in AFN, $\bar{x}$ tends towards $-\infty$, leading to gradient explosion. To solve this problem, we bound the gradient to successfully solve this problem. More details please see in Section \ref{sec:result}.

\subsection{Differences between AFN and ASRNorm}

From our point of view, IN may struggle to capture the inter-instance information within a batch, which can hinder the learning of relationships between different images. Motivated by ASRNorm, we introduce additional parameters to BN, enhancing the generalisation capability of neural networks. This modification results in a more stable training process compared to ASRNorm which will be
studied in the experiments presented in the next section.

%% file: sections/exp.tex
\section{Experiments}

\subsection{Experimental Settings}
\label{sec:setting}

\textbf{Datasets.} We conduct experiments on four standard benchmarks for image classification~\cite{Krizhevsky2009LearningML, Ganin2014UnsupervisedDA, Netzer2011ReadingDI}, including \emph{CIFAR-10, CIFAR-100, SVHN, MNIST-M}, and three standard benchmarks for domain generalisation~\cite{Volpi2019AddressingMV, Xiao2022LearningTG, Volpi2018GeneralizingTU, Zhao2020MaximumEntropyAD}, including \emph{Digits, CIFAR-10-C and PACS}.

\begin{figure}[t]
    \centering
    \includegraphics[width=1\linewidth]{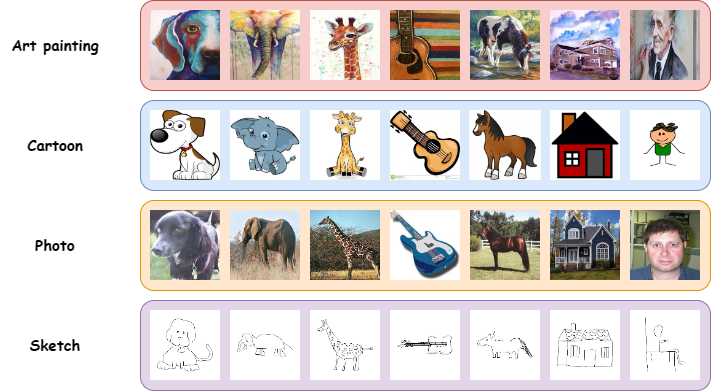}
    \caption{Illustration of domain generalisation with the PACS benchmark. Single-domain generalisation aims at training a model on one source domain data, while generalising well to other domains with very different visual presentations. For multi-domain generalisation, one source domain data is used for training, and the others are used for testing.}
    \label{pacs}
\end{figure}

(1) \emph{CIFAR-10, CIFAR-100}: This benchmark consists of 60,000 32x32 RGB images in 10 different classes, with 6,000 images per class. CIFAR-100 dataset is an extension of CIFAR-10 and contains 100 classes, with 600 images per class. Each image is 32x32 color image.

(2) \emph{Digits}: This benchmark consists of four digits datasets: MNIST~\cite{LeCun1989BackpropagationAT}, SVHN~\cite{Netzer2011ReadingDI}, MNIST-M~\cite{Ganin2014UnsupervisedDA}, USPS~\cite{2002A}. Images in MNIST and USPS are grey images, but in SVHN and MNIST-M are color images. In the domain generalisation task, to make four datasets have compatible shapes, all images are resized to 32 x 32 pixels, and the channels of the datasets are verified to be the same. We use one dataset as a training dataset, and the rest of them as a testing dataset.

(3) \emph{CIFAR-10-C}~\cite{Hendrycks2018BenchmarkingNN}: This benchmark is proposed to evaluate the robustness of 109 types of corruptions with 5 levels of intensities. The original CIFAR-10 is used for training and corruptions are only applied to the testing images. The corruption intensity can measure the level of domain discrepancy.  

(4) \emph{PACS}~\cite{Li2017DeeperBA}: Figure~\ref{pacs} shows the benchmark for domain generalisation containing four domains: \emph{art paint, cartoon, sketch}, and \emph{photo}, which share the same categories that include \emph{dog, elephant, giraffe, guitar, house, horse and person}. To align the results in the previous work, for this dataset, we consider the same settings: 1) training a model with one domain data and testing with the rest three; 2) training a model on three domains and testing on the remaining one. In both settings, we remove the domain label and mix the data from different domains.

\textbf{Experiments Details.} It is noted that $C_{stan}, C_{rescale}$ are set to $C/2, C/16$, respectively. The initialisation of hyperparameters follows the settings of ASRNorm~\cite{ASRNorm}.

(1) For \emph{Digits}. In the domain generalisation task, we use the ConvNet architecture~\cite{LeCun1989BackpropagationAT} \emph{(conv-pool-conv-pool-fc-fc-softmax)} with ReLU following each convolution. Since this model has no normalisation layer, AFN is inserted after each convolution layer before ReLU. We follow this experimental setting: using Adam~\cite{Kingma2014AdamAM} optimiser with a learning rate $10^{-3}$ and training batch size 128, testing batch size 256. In the image classification task, we conduct experiments on various backbones, by using SGD as the optimiser with the learning rate initilised with $0.1$. 

(2) For \emph{CIFAR-10-C}, we use WRN-40-4~\cite{Zagoruyko2016WideRN}, SGD with Nesterov momentum(0.9), and a batch size of 128. The initial learning rate is 0.1, annealing with ALRS scheduler~\cite{Chen2022BootstrapGA}. 

(3) For \emph{PACS}, we use ResNet-18 pretrained on ImageNet, the Adam optimiser with an initial learning rate $10^{-4}$, ALRS as a learning rate scheduler with training epochs 50, and mini-batch size is 32. The model learned AFN with the RSC~\cite{Huang2020SelfChallengingIC} procedure, which is an algorithm to virtually augment challenging data by shutting down the dominant neurons that have the largest gradients during training.

(4) For CIFAR-10, CIFAR-100 datasets, we follow the same experimental setting: Adam optimiser with an initial learning rate $10^{-3}$, scheduled by ARLS. The model is trained with a mini-batch size 128 for about 200 epochs.

\subsection{Main results and analysis}
\label{sec:result}

Our approach outperforms the previous SOTA normalisation method~(ASRNorm) on single domain generalisation tasks by \textbf{$0.9\%, 0.6\%, 1.3\%, 1.6\%$} on the Digits(two experiments), CIFAR-10-C, and PACS benchmarks, respectively.

\begin{table}[!ht]
    \centering
    \scalebox{0.8}{
    \begin{tabular}{ccccc}
    \hline
        Method & SVHN & USPS & MNIST-M & Avg. \\ \hline
        BN & 27.8 & 76.9 & 52.7 & 52.5 \\ 
        ASRNorm & 34.1 & 78.5 & 64.3 & 59.0 \\ 
        AFN(our) & 33.8 & \textbf{81.2} & 63.8 & \textbf{59.6} \\ \hline
        $\Delta$ w.r.t ASRNorm & -0.3 & 2.7 & -0.5 & 0.6
    \end{tabular}
    }
    \caption{Single domain generalisation accuracies on Digits. Choose MNIST as the training set, and the rest of the domains as the testing set, increasing the channels in USPS and MNIST to make the datasets compatible.}
    \label{tab:mnist}
\end{table}

\begin{table}[!ht]
    \centering
    \scalebox{0.8}{
    \begin{tabular}{cccccc}
    \hline
        Method & MNIST & SVHN & USPS & MNIST-M & Avg. \\ \hline
        BN & 74.0 & 30.3 & 73.2 & 38.7 & 54.1 \\ 
        BIN & 71.4 & 30.6 & 70.6 & 42.5 & 53.8 \\ 
        ASRNorm & 75.6 & \textbf{34.0} & 70.9 & \textbf{45.5} & 56.5 \\ \hline
        AFN(our) & \textbf{77.6} & 33.8 & \textbf{73.4} & 44.8 & \textbf{57.4} \\ 
        $\Delta$ w.r.t ASRNorm & 2.0 & -0.2 & 2.5 & -0.7 & 0.9
    \end{tabular}
    }
    \caption{Single domain generalisation accuracies on Digits. Choose one domain as the training set, and the results on the remaining domains as the testing set, reducing the channels in SVHN and MNIST-M to make the datasets compatible.}
    \label{tab:digits}
\end{table}

\textbf{Results on Digits}: Table \ref{tab:mnist} and \ref{tab:digits} show the results on the \emph{Digits} benchmark. The proposed AFN is compared with the previous SOTA normalisation method ASRNorm. Our method outperforms both the baseline and the SOTA methods on average.

\begin{table}[!ht]
\scriptsize
    \centering
    \scalebox{1}{
    \begin{tabular}{ccccccc}
    \hline
        Method & Level 1 & Level 2 & Level 3 & Level 4 & Level 5 & Avg. \\ \hline
        BN & 87.8 & 81.5 & 73.2 & 75.5 & 56.1 & 74.8 \\ 
        ASRNorm & \textbf{89.4} & 86.1 & 82.9 & 78.6 & 72.9 & 82.0 \\ \hline
        AFN(our) & 89.3 & \textbf{86.6} & \textbf{83.7} & \textbf{79.9} & \textbf{77.0} & \textbf{83.3} \\ 
        $\Delta$ w.r.t ASRNorm & -0.1 & 0.5 & 0.8 & 1.3 & 4.1 & 1.3
    \end{tabular}
    }
    \caption{Single domain generalisation accuracies on CIFAR10-C. CIFAR-10 is used as the training domain, while CIFAR-10-C with different corruption types and corruption levels is used as the testing domain.}
    \label{tab:cifar10c}
\end{table}

\textbf{Results on CIFAR-10-C}: We show average accuracies of the 19 corruption types for each level of intensity of CIFAR-10-C in Table \ref{tab:cifar10c}. Similar to the results on the Digits benchmark, AFN achieves larger improvements on the more challenging domains, because the early stage of training, behaving like BN, helps collect the information from different samples. From Figure \ref{cifar10c}, it is obvious that our method has better generalization ability with respect to other methods, with the increase of corruption levels. 

\begin{table}[!ht]
    \centering
    \scalebox{0.8}{
    \begin{tabular}{cccccc}
    \hline
        Method & Art painting & Cartoon & Sketch & Photo & Avg. \\ \hline
        ERM & 70.9 & 76.5 & 53.1 & 42.2 & 60.7 \\ 
        RSC & 73.4 & 75.9 & 56.2 & 41.6 & 61.8 \\ 
        RSC+ASRNorm & 76.7 & \textbf{79.3} & \textbf{61.6} & 54.6 & 68.1 \\ \hline
        RSC+AFN(our) & \textbf{77.3} & 78.2 & 61.3 & \textbf{62.1} & \textbf{69.7} \\ 
        $\Delta$ w.r.t RSC+AFN & 0.6 & -1.1 & -0.3 & 7.5 & 1.6
    \end{tabular}
    }
    \caption{Single domain generalisation accuracies on PACS. One domain is used as the training set and the other domains are used as the testing set.}
    \label{tab:spacs}
\end{table}

\textbf{Results on PACS}: In table \ref{tab:spacs}, we show the results on PACS where we use one domain for training and the remaining three for testing. Our method improves the performance of RSC in the Art painting and Photo domains, and reaches better average accuracy on this task. Table \ref{tab:mpacs} shows the results on PACS for the multi-source domain setting. Similarly, AFN slightly outperforms the previous SOTA method.

\begin{table}[!ht]
    \centering
    \scalebox{0.7}{
    \begin{tabular}{cccccc}
    \hline
        Method & Art painting & Cartoon & Sketch & Photo & Avg. \\ \hline
        ERM & 82.7 & 78.7 & 78.6 & 95.1 & 83.8 \\ 
        RSC & 82.7 & 79.8 & 80.3 & 95.6 & 84.6 \\ 
        RSC+ASRNorm & \textbf{84.8} & 81.8 & 82.6 & \textbf{96.1} & 86.3 \\ \hline
        RSC+AFN(our) & 84.4 & \textbf{82.4} & \textbf{82.9} & \textbf{96.1} & \textbf{86.5} \\ 
        $\Delta$ w.r.t RSC+AFN & -0.4 & 0.6 & 0.3 & 0 & 0.2
    \end{tabular}
    }
    \caption{Multi-domain generalisation accuracies on PACS. One domain is used as the test set and the other domains are used as the training sets. During training, we remove the domain labels.}
    \label{tab:mpacs}
\end{table}

\begin{table}[!ht]
\scriptsize
    \centering
    \scalebox{1}{
    \begin{tabular}{c|cc|c}
    \hline
        Backbone & BN & ASRNorm & AFN \\ \hline
        Resnet18 & 0.9427 & 0.9441 & \textbf{0.9462} \\ 
        Renset32 & 0.9511 & 0.9472 & \textbf{0.9538} \\ 
        WRN\_16\_2 & 0.9512 & 0.9489 & \textbf{0.9523} \\
        VGG13 & 0.9411 & 0.9409 & \textbf{0.9427} \\
        VGG16 & 0.9433 & nan & \textbf{0.9461} \\ 
    \end{tabular}
    }
    \caption{Image classification accuracies on SVHN.}
    \label{tab:svhn}
\end{table}

\begin{table}[!ht]
\scriptsize
    \centering
    \scalebox{1}{
    \begin{tabular}{c|cc|c}
    \hline
        Backbone & BN & ASRNorm & AFN \\ \hline
        Resnet18 & 0.9880 & 0.9870 & \textbf{0.9887} \\ 
        Renset32 & 0.9887 & 0.9847 & \textbf{0.9893} \\  
        WRN\_16\_2 & 0.9859 & 0.9855 & \textbf{0.9891} \\
        VGG13 & 0.9830 & 0.9831 & \textbf{0.9875} \\
        VGG16 & 0.9838 & nan & \textbf{0.9849} \\ 
    \end{tabular}
    }
    \caption{Image classification accuracies on MNIST-M.}
    \label{tab:mnist-m}
\end{table}

\begin{table}[!ht]
\scriptsize
    \centering
    \scalebox{1}{
    \begin{tabular}{c|cc|c}
    \hline
        Backbone & BN & ASRNorm & AFN \\ \hline
        Resnet18 & 0.9495 & 0.9400 & \textbf{0.9515} \\ 
        Renset32 & 0.9419 & 0.9308 & \textbf{0.9432} \\ 
        Resnet56 & 0.9450 & 0.9388 & \textbf{0.9472} \\ 
        WRN\_16\_2 & 0.9404 & 0.9353 & \textbf{0.9447} \\
        VGG13 & 0.9179 & 0.9238 & \textbf{0.9343} \\
        VGG16 & 0.9435 & 0.9244 & \textbf{0.9448} \\ 
        VGG19 & 0.9442 & 0.9334 & \textbf{0.9451} \\ 
    \end{tabular}
    }
    \caption{Image classification accuracies on CIFAR-10.}
    \label{tab:cifar10}
\end{table}

\begin{table}[!ht]
\scriptsize
    \centering
    \scalebox{1}{
    \begin{tabular}{c|cc|c}
    \hline
        Backbone & BN & ASRNorm & AFN \\ \hline
        Resnet18 & 0.7327 & 0.6418 & \textbf{0.7457} \\ 
        Renset34 & 0.7521 & 0.7054 & \textbf{0.7569} \\ 
        Resnet50 & 0.7509 & 0.6830 & \textbf{0.7558}\\ 
        WRN\_40\_10 & 0.7875 & 0.7538 & \textbf{0.7956} \\ 
        VGG13 & 0.6995 & 0.5150 & \textbf{0.7032} \\ 
        VGG16 & 0.6922 & \text{nan} & \textbf{0.6972} \\  
    \end{tabular}
    }
    \caption{Image classification accuracies on CIFAR-100. }
    \label{tab:cifar100}
\end{table}

\textbf{Results on image classification.} Table \ref{tab:svhn} and \ref{tab:mnist-m} show the results on SVHN and MNIST-M. Our method surpasses BN and ASRNorm, also we see there exist nan values in the tables, which mean the inability to keep the gradients steady during training, leading to training failure. However, our method is capable of keeping the gradient steady to finish our training.

Table \ref{tab:cifar10} and \ref{tab:cifar100} shows the results on CIFAR-10, CIFAR-100. Our method can be successfully applied to image classification tasks and outperform the original BN, which means our method can replace BN in the network to improve performance. It is noted that the previous SOTA method cannot be applied to this task, and may even deteriorate the network performance. ASRNorm is not suitable for image classification tasks due to its inability to consider mini-batch information and address covariance shift effectively, in the early training stage. In contrast, our method leverages Batch Normalisation statistics to capture mini-batch information, making it more suitable. Additionally, our method demonstrates superior stability during training, reducing the risk of gradient explosion associated with ASRNorm.

%% file: sections/conclusion.tex
\section{Conclusion}
We designed a new normalisation function, Adaptive Fusion Normalisation by adding more parameters to the Batch Normalisation function model, which outperforms previous normalisation methods across specific tasks. Extensive experimentation was conducted to demonstrate the advantages of our approach, showcasing the enhanced generalisation capabilities of our model. In the future, we will extend our approach to other fields like speech recognition. Also, we will explore the architectures of the encoder and decoder to get better results.